\documentclass[10pt,twocolumn,letterpaper]{article}

\usepackage[utf8]{inputenc}
\usepackage[T1]{fontenc}
\usepackage[margin=0.75in,columnsep=0.25in]{geometry}
\usepackage{amsmath,amssymb}
\usepackage{newtxtext,newtxmath}
\usepackage{microtype}
\usepackage[numbers,sort&compress]{natbib}
\usepackage{graphicx}
\usepackage{booktabs}
\usepackage{array}
\usepackage{multirow}
\usepackage[font=small,labelfont=bf]{caption}
\usepackage{subfig}
\usepackage{xurl}
\usepackage[hidelinks]{hyperref}
\hypersetup{
  pdftitle={How to Better Train VLAs: Lessons Learned From the REAL-I Challenge at ICRA 2026},
  pdfauthor={Jiaming Wang; Jizhuo Chen; Diwen Liu; Wang Song; Qiang Wang; Jie Ren; Chao Fu; Dingkun Zhu; Minchi Ruan; Hongtong Li; Yuhua Jiang; Zhiwei Xue; Yongping Pan; Harold Soh}
}
\newcommand{\reali}{REAL-I}

\title{How to Better Train VLAs: Lessons Learned From the \reali{} Challenge at ICRA 2026}
\author{%
  \normalsize Jiaming Wang\textsuperscript{*}, Jizhuo Chen, Diwen Liu, Wang Song, Qiang Wang,\\[0.3em]
  \normalsize Jie Ren, Chao Fu, Dingkun Zhu, Minchi Ruan, Hongtong Li,\\[0.3em]
  \normalsize Yuhua Jiang, Zhiwei Xue, Yongping Pan, Harold Soh\\[0.6em]
  \small \textsuperscript{*}Corresponding author: jiaming.w@u.nus.edu.
}
\date{}

\begin{document}
\twocolumn[\begin{@twocolumnfalse}
\maketitle
\begin{abstract}
How can robot policies learn more effectively from a fixed demonstration budget? The first Real-world Embodied AI Learning (\reali{}) Challenge at ICRA 2026 examined this question through simulation, real-robot evaluation, and an on-site final on a shared dual-arm humanoid platform. We describe the challenge tasks, data and deployment interfaces, and competition results, then compare the approaches contributed by NUS-CLEAR, RCL-Lab, and Deeptouch.ai. Their systems combined pretrained vision-language-action models and task-specific imitation policies with different strategies for data curation, staged adaptation, checkpoint selection, and action-space design. The team reports highlight the importance of adapting to the deployment environment while retaining prior capabilities, treating demonstration quality at an appropriate temporal scale, and suppressing errors in inactive robot components. They also expose the limitations of offline action-prediction metrics for forecasting closed-loop success. These observations motivate a view of fixed-data robot learning that integrates data, adaptation, evaluation, and deployment.
\end{abstract}
\vspace{1em}
\end{@twocolumnfalse}]

\section{Introduction}

Robot learning can improve along two complementary axes: scaling data and learning more effectively from a given dataset. Although additional demonstrations can often be collected, expert teleoperation and real-robot interaction are costly. This motivates a basic question: for a fixed data budget, how can we learn the strongest possible policy? Studying this regime isolates improvements in learning strategy from gains due simply to more data.

Given the same trajectories, robot-learning systems can use them in different ways. \emph{Modular approaches} decompose perception, reasoning, planning, and control, providing interpretable interfaces and reusable structure \cite{wang2023polo,wang2025genie,ichter2023saycan,huang2023voxposer, wang2025taskgrasp}. \emph{End-to-end approaches} instead learn observation-to-action mappings and can exploit shared representations, multimodal action distributions, and large-scale pretraining \cite{chi2023diffusionpolicy,physicalintelligence2025pi05, nvidia2026gr00tn17}. Hybrid systems combine both. Yet regardless of system organization, learning from a fixed dataset is fundamentally constrained by how well its trajectories cover the states encountered during closed-loop execution.

This coverage constraint is especially consequential in real-world embodied AI. Small closed-loop errors can move a behavior-cloned policy beyond the demonstrated state distribution, while contact, occlusion, and multimodal behavior further amplify this gap in manipulation \cite{mandlekar2021robomimic}. Interactive correction methods such as DAgger address such distribution shift when additional labeled interaction is available, whereas offline reinforcement learning must avoid extrapolating to unsupported actions \cite{ross2011dagger,levine2020offline}. The \reali{} Challenge focuses on the complementary fixed-data regime, using standardized policy training and physical evaluation on a common humanoid platform.

\section{The \reali{} Challenge}

\begin{figure*}[t]
    \centering
    \includegraphics[width=0.98\textwidth,height=0.25\textheight,keepaspectratio]{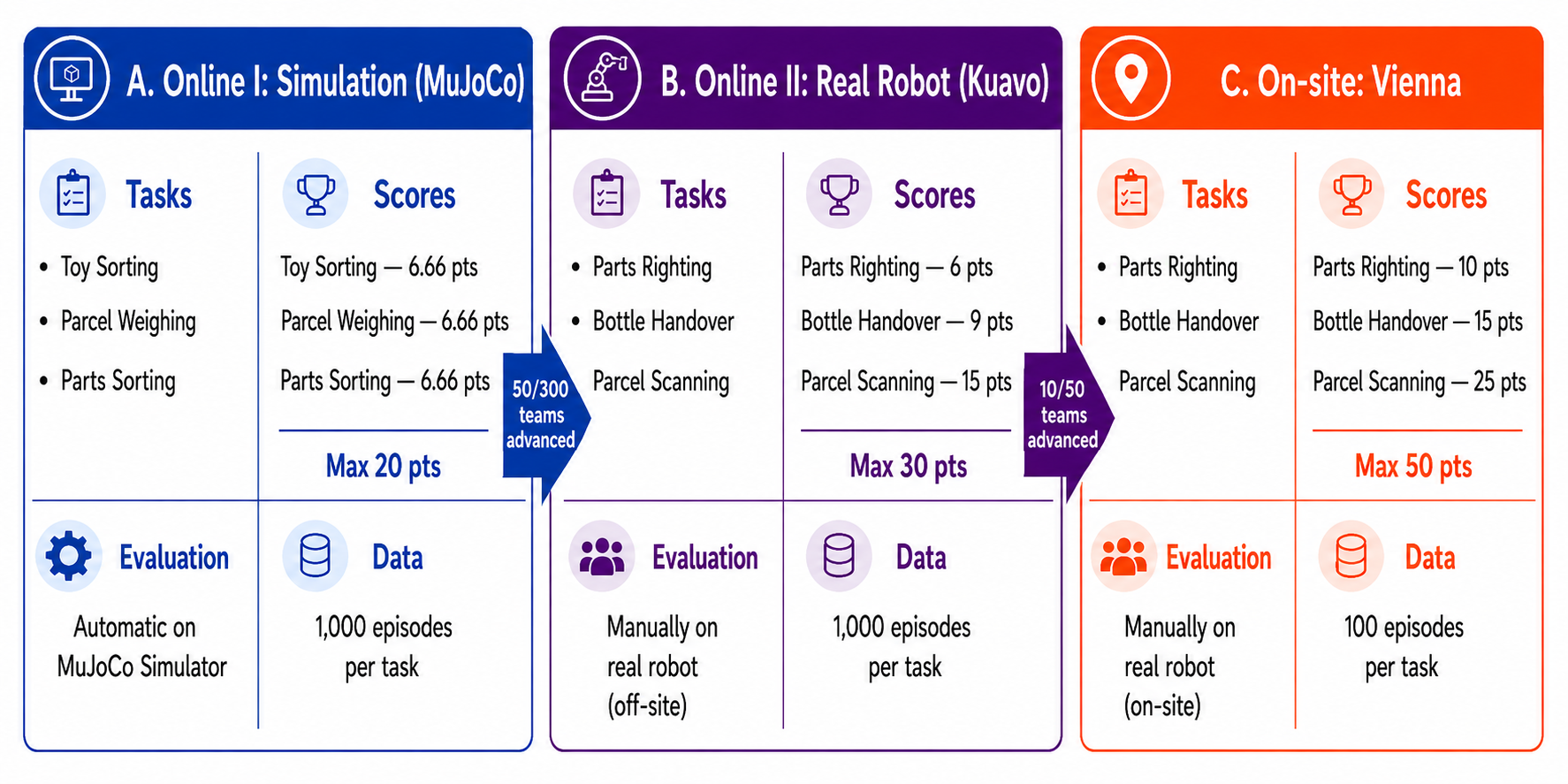}
    \caption{Overview of the three \reali{} competition stages, including tasks, scoring weights, evaluation settings, dataset sizes, and team progression.}
    \label{fig:competition_overview}
\end{figure*}

The first Real-world Embodied AI Learning Challenge (\reali{}) was organized at the 2026 IEEE International Conference on Robotics and Automation (ICRA) to address the need for large-scale data, standardized evaluation, and direct access to physical systems. It provided open access to real robots with unified benchmarking and evaluation tools, and a large-scale industrial dataset collected on homogeneous dual-arm humanoid robots \cite{realichallenge2026}. Crucially, \reali{} introduced a constraint designed to reflect the practical scarcity of data in real-world robotics: participants were prohibited from collecting or using any additional datasets beyond those provided by the challenge organizers. This restriction places greater emphasis on data-efficient learning and better reflects deployment settings in which acquiring new robotic data is costly and time-consuming. Figure~\ref{fig:competition_overview} summarizes the three-stage competition.

\begin{figure*}[t]
    \centering
    \subfloat[Daily Chemical Bottle Pick-and-Place.]{%
        \includegraphics[width=0.32\textwidth]{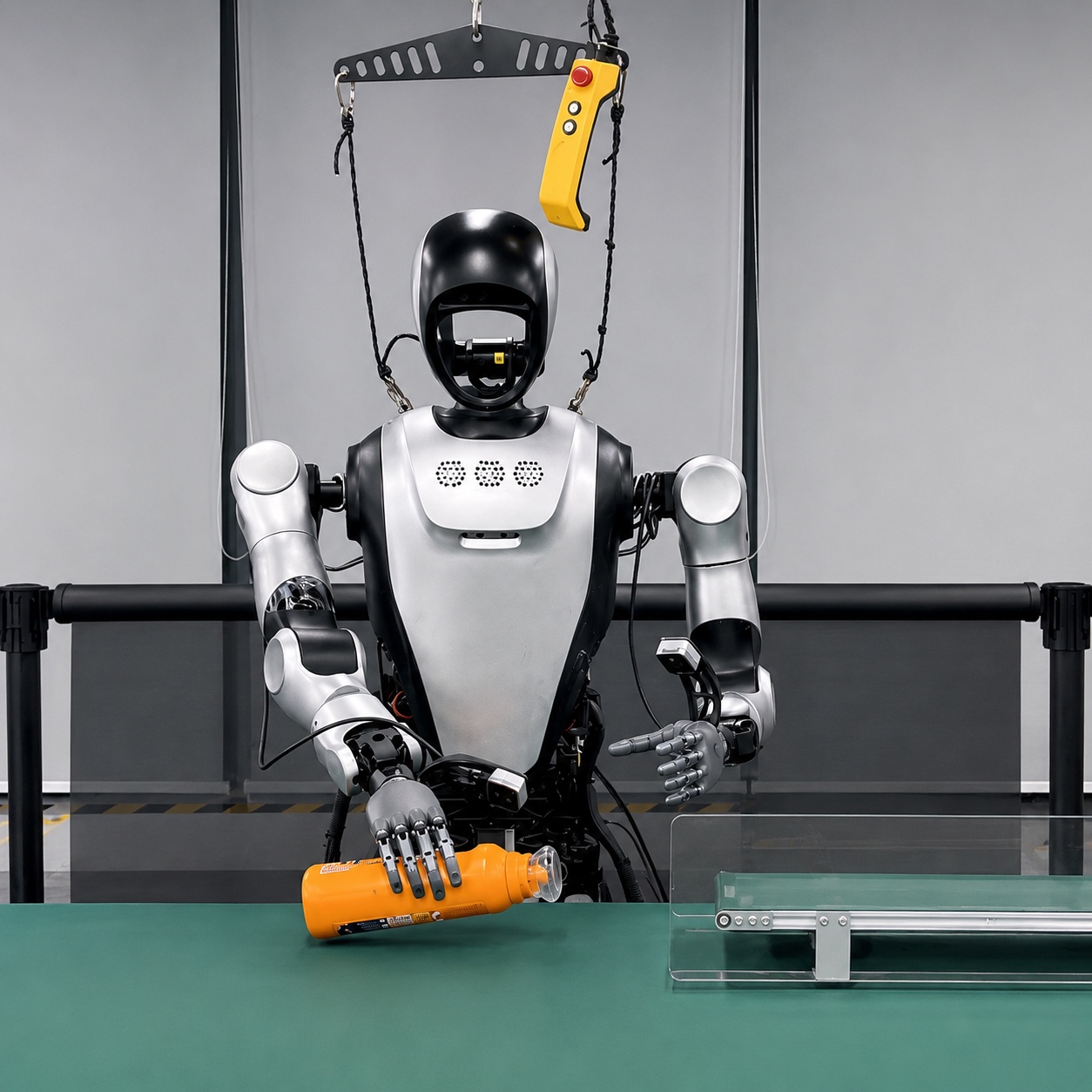}%
        \label{fig:task_bottle}}
    \hfill
    \subfloat[Express Package Scanning.]{%
        \includegraphics[width=0.32\textwidth]{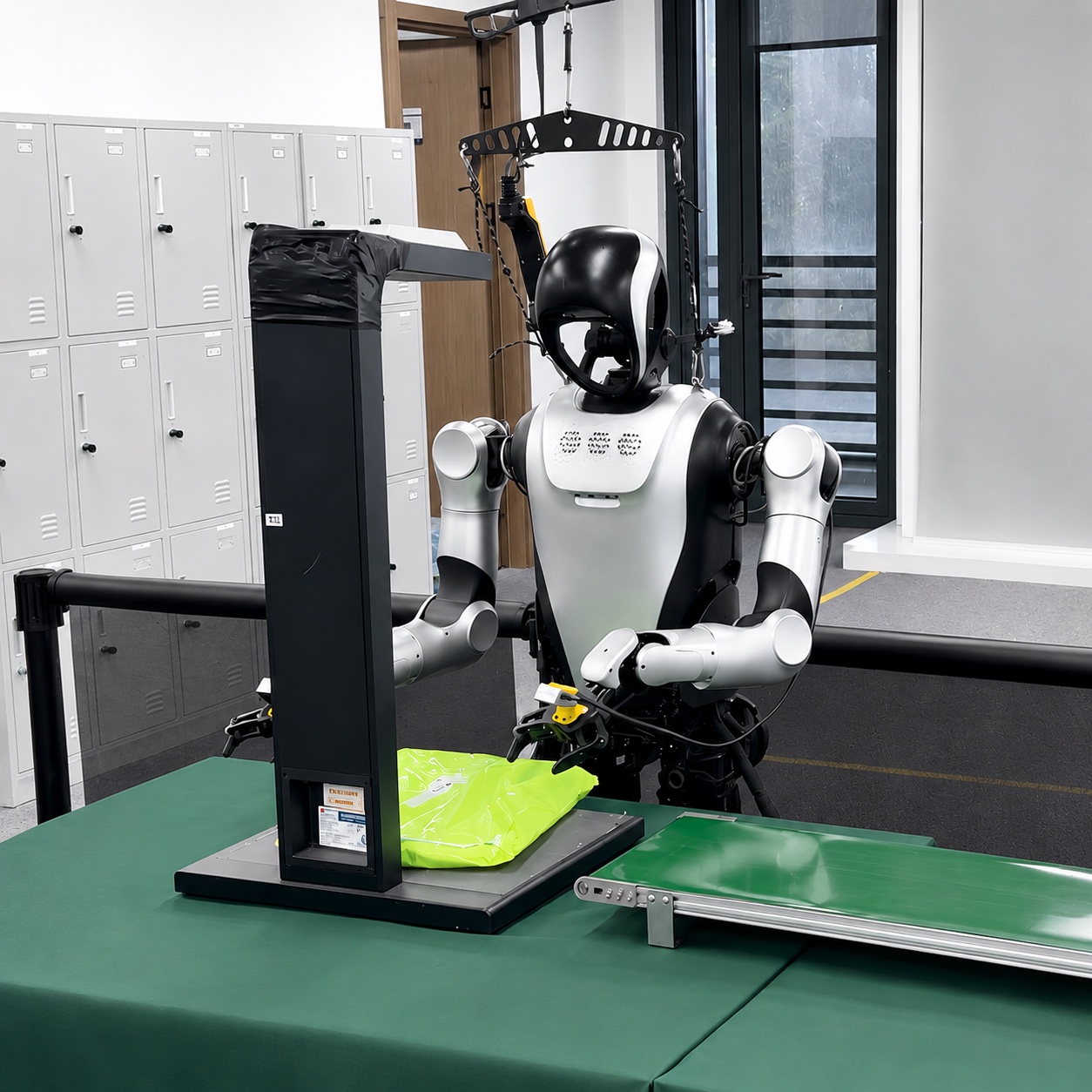}%
        \label{fig:task_package}}
    \hfill
    \subfloat[Metal Parts Righting.]{%
        \includegraphics[width=0.32\textwidth]{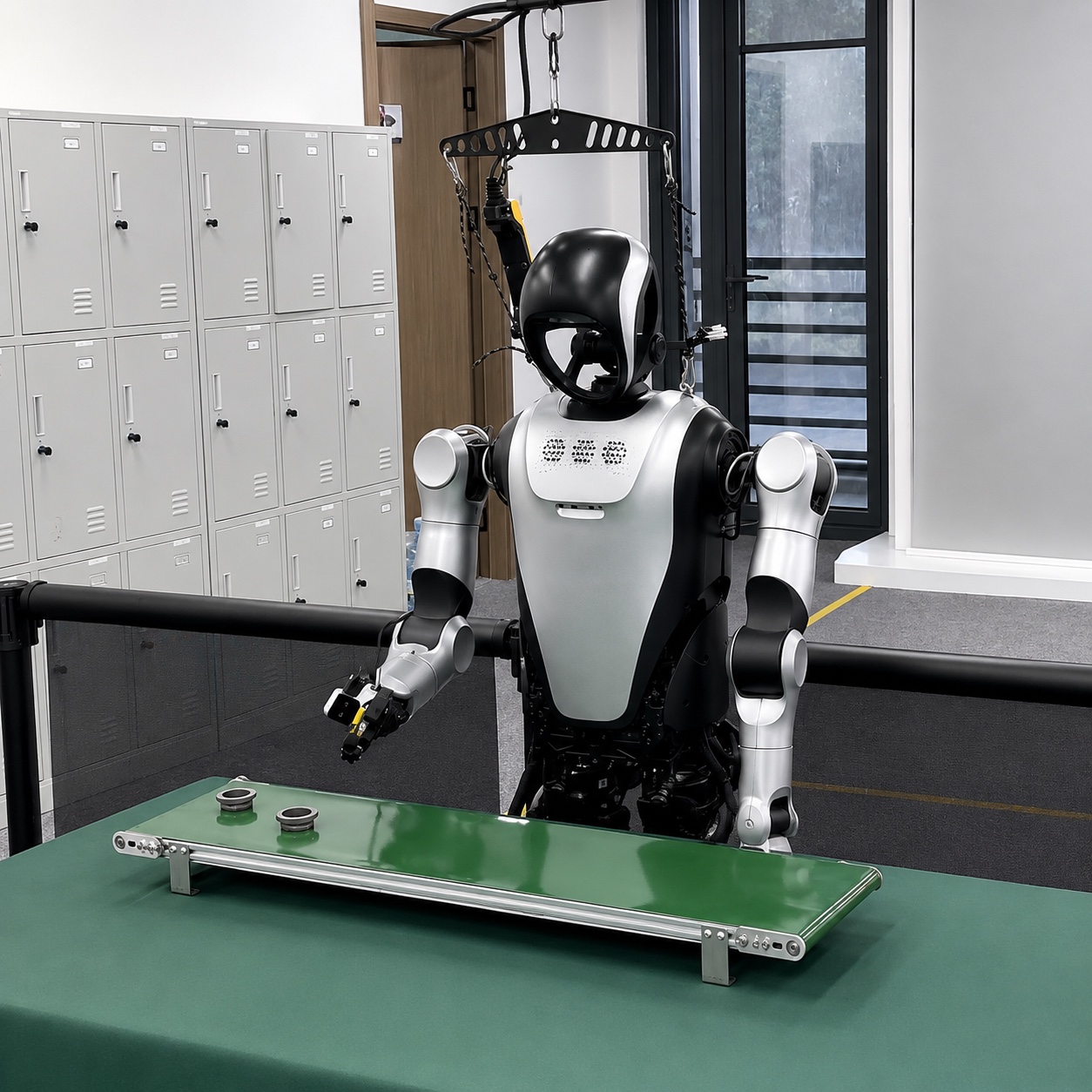}%
        \label{fig:task_metal}}
    \caption{Robot and workcell setups for the three on-site final tasks: Daily Chemical Bottle Pick-and-Place, Express Package Scanning, and Metal Parts Righting. Photographs \textcopyright{} 2026 Leju (Shenzhen) Robotics Co., Ltd. Used with permission.}
    \label{fig:onsite_tasks}
\end{figure*}

\subsection{Competition Stages and Tasks}

The competition used a progressive format covering the simulation stage, the real-robot stage, and the ICRA 2026 on-site final~\cite{realichallenge2026}. This staged design was intended to evaluate not only whether a policy could fit the released demonstrations, but also whether the complete offline-to-real pipeline could support stable deployment on physical humanoid robots. Across stages, participants used the same data-processing, training, and inference workflow, while robot execution was mediated by the Kuavo robot-control stack.

The simulation stage introduced participants to the Kuavo data format, baseline code, and deployment workflow through Toy Sorting, Parcel Weighing, and Conveyor Belt Parts Sorting. This stage provided a controlled setting for checking data conversion, observation construction, action representation, policy loading, and closed-loop rollout before physical evaluation. It also reduced avoidable failures caused by incorrect preprocessing or mismatched policy interfaces.

The real-robot stage evaluated policies on physical robots through Metal Parts Righting, Daily Chemical Bottle Pick-and-Place, and Express Package Scanning. This stage tested whether policies trained from fixed demonstrations could remain stable under real sensing, contact, calibration, and timing constraints. Compared with simulation, failures at this stage could arise from perceptual latency, grasp instability, object pose variation, end-effector timing, and discontinuities in the predicted action sequence. Representative robot setups for these physical tasks are shown in Fig.~\ref{fig:onsite_tasks}.

The on-site final used the same three tasks, summarized in Table~\ref{tab:onsitetasks}, but in different physical setups. Although no distribution shift was intentionally introduced, natural variations in lighting, robot configuration, table layout, and object poses arose from deployment at a new location. This closely reflects practical deployment, where policies trained in one environment must transfer to a client's site under similar variations. Participants could use all organizer-provided datasets released in previous stages when preparing for the final; no additional task data could be collected by the teams.

Overall, the staged format made REAL-I a benchmark for deployment-ready fixed-data robot learning. The simulation stage checked whether teams could use the released software and data correctly; the real-robot stage exposed the gap between offline validation and hardware execution; and the on-site final measured robustness under time-limited industrial manipulation tasks. The final comparison therefore depended on the learned policy as well as data preprocessing, action representation, model selection, deployment control, and interaction with the robot control system.

\begin{table*}[t]
\caption{On-site \reali{} scenes and per-submission scoring criteria at ICRA 2026 \cite{realichallenge2026}.}
\label{tab:onsitetasks}
\centering
\begin{tabular}{p{0.20\textwidth}p{0.33\textwidth}p{0.39\textwidth}}
\toprule
Scene & Task & Scoring criteria \\
\midrule
Metal Parts Righting & Flip small metal parts from face-down to face-up on a conveyor belt. & 2.5 points per part successfully grasped and flipped, plus a 2.5-point all-parts bonus (10 points maximum; 3 min). \\
Daily Chemical Bottle Pick-and-Place & Grasp a bottle, perform an in-air bimanual handover, and place it on a conveyor belt. & 3 points for grasping, 6 for bimanual handover, and 6 for conveyor placement (15 points maximum; 3 min). \\
Express Package Scanning & Place two parcels on a scanning table, correct label orientation when needed, and transfer them to a conveyor belt. & 2.5 points per scanner placement, 5 for correcting a label-down parcel, and 10 per conveyor placement, capped at 25 points (6 min). \\
\bottomrule
\end{tabular}
\end{table*}

\subsection{Data, Software, and Evaluation Pipeline}
The REAL-I software stack was organized as an offline-to-real pipeline rather than as an isolated policy-training benchmark. All dataset conversion, policy training, simulation evaluation, and real-robot inference were performed through the Kuavo learning repository, while low-level robot execution was handled by the Kuavo ROS control stack. This separation made the learning interface explicit: the learning system consumed synchronized visual and proprioceptive observations and produced robot-level action targets, whereas the control stack was responsible for translating those targets into physically executable whole-body motion.

The released demonstrations were stored as ROS bags and converted into the LeRobot dataset format through the Kuavo learning pipeline. The official competition materials describe the dataset as large-scale industrial robot data collected on homogeneous dual-arm humanoid robots, including synchronized multi-view RGB/depth videos, robot joint states, end-effector states, and control commands \cite{realichallenge2026}. In the baseline workflow, these raw streams were mapped into a unified observation-action schema for policy training and deployment. The conversion process handled timestamp alignment, camera resizing, platform-dependent joint indexing, arm selection, and action normalization, so that policies trained offline could be evaluated through the same inference-side interface in simulation and on the real robot.

\begin{figure*}[t]
    \centering
    \includegraphics[width=0.98\textwidth]{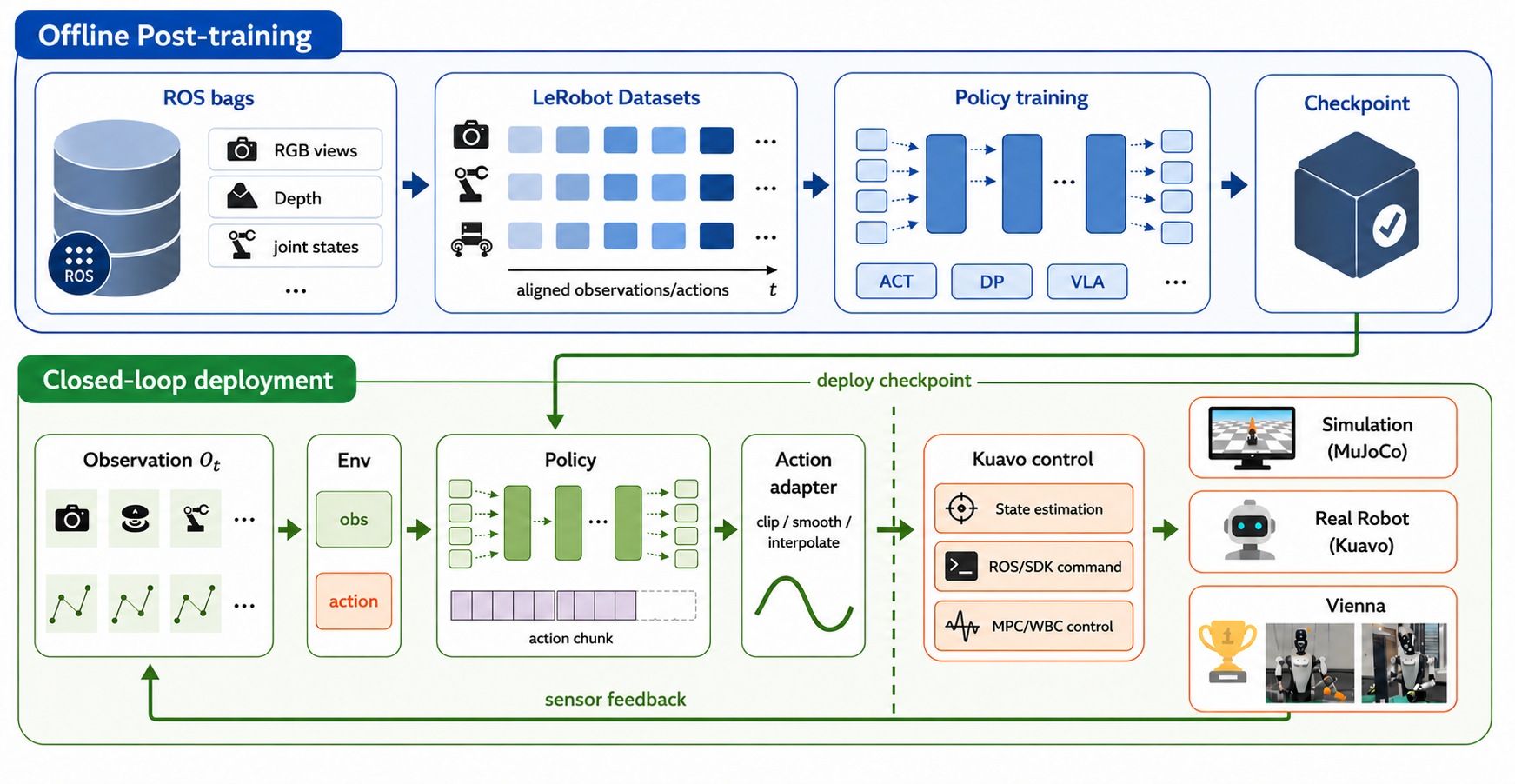}
    \caption{Overview of the REAL-I post-training and deployment pipeline. Demonstration data are converted from ROS bags into the LeRobot dataset format for policy training. During deployment, the trained checkpoint is executed through an environment wrapper that converts synchronized observations into policy inputs and maps predicted action chunks to robot commands. The robot-control stack handles state estimation, ROS/SDK command interfaces, and MPC/WBC-based control before sending executable commands to the Kuavo robot.}
    \label{fig:post_training_pipeline}
\end{figure*}

At deployment time, the trained policy was wrapped by a Gym-style Kuavo environment. The environment provided the same logical interface in simulation and on the real robot: it collected the latest aligned observations, constructed the policy input dictionary, invoked the policy, post-processed the predicted action, and executed one environment step. In the joint-control setting used by the baseline pipeline, each arm action consisted of seven arm joint-position targets plus one scalar end-effector command. A dual-arm policy therefore outputs a 16-dimensional action vector, corresponding to the left-arm joints, the left end effector, the right-arm joints, and the right end effector. Single-arm tasks used the same convention but masked or filled the inactive arm so that the robot interface remained well defined.

The deployment environment also served as the boundary between learned policy outputs and robot-safe command execution. Predicted actions were clipped to configured joint and end-effector limits before execution. When direct whole-body-control execution was enabled, the environment interpolated policy-rate commands to a higher control rate and applied low-pass filtering before sending arm targets. In the default configuration, policy inference ran at a lower rate, while interpolated control commands were issued at the robot-control rate. This design reduced discontinuities caused by action chunking or slow neural-network inference and made the policy output compatible with the timing assumptions of the motion-control stack.

Communication with the robot was implemented through the Kuavo SDK and ROS. Before policy execution, the robot arm was switched into an external-control mode. The deployment environment then sent arm-joint targets and end-effector commands through the robot-control interface, while receiving synchronized camera, joint-state, and end-effector feedback from the same ROS-based system. The simulator and the physical robot exposed the same learning-side interface, which made the simulation and real-robot stages structurally comparable.

Below this learning-facing interface, the Kuavo control stack maintained whole-body feasibility. The motion-control system used ROS nodes for state estimation, target handling, model predictive control, and whole-body control. High-level targets from the learning system were treated as desired arm or whole-body references; the controller then generated executable joint commands while considering the current robot state, balance, joint limits, and the active control mode. For mobile-manipulation functions, the control stack exposed OCS2-based MPC target trajectories for torso and arm motion, while the WBC layer converted planned motion and state estimates into joint-level commands. In the implementation, the WBC module was instantiated as a weighted whole-body controller and solved as a quadratic program.
The cost matrices are assembled from weighted tracking tasks, including center-of-mass, base motion, contact-force, stance/swing-leg, and arm-joint acceleration tasks. The constraints include the floating-base equations of motion, joint-torque limits, and friction-cone constraints. The optimized solution provides planned body acceleration, joint acceleration, contact forces, and joint torques, from which executable joint-level commands are generated. Thus, the neural policy did not directly command motor torques. Instead, it produced task-level joint-position setpoints and end-effector commands that were mediated by the ROS control stack before reaching the hardware.

This architecture was important for the competition because real-robot performance depended on the complete path from data to execution. A policy trained on fixed demonstrations could fail not only because of model error, but also because of mismatched observation timing, action scaling, inactive-arm drift, command-rate mismatch, or discontinuities between predicted action chunks. The shared pipeline reduced these sources of accidental variation by standardizing data conversion, observation construction, policy loading, action post-processing, ROS communication, and robot-side control. At the same time, it left room for teams to improve data filtering, model architecture, action representation, checkpoint selection, and deployment-time smoothing. Figure~\ref{fig:post_training_pipeline} summarizes this data-to-deployment pipeline.

Beyond this robot-side execution path, the competition workflow treated policy learning, software packaging, evaluation, and feedback as one iterative system. Teams started from local or cloud ROS-bag datasets containing RGB images, optional depth, robot state, and action streams. These logs were converted into a standardized LeRobot dataset for policy training. Trained model weights, configuration files, and offline dependencies were then packaged into a Docker image for evaluation. The same interface supported the simulation stage, the real-robot stage on Kuavo, and the on-site final, so that differences across stages reflected task difficulty and physical conditions rather than a different deployment contract. Each stage returned feedback, including quantitative scores, leaderboard position, execution videos, system logs, diagnostics, and qualitative judging, closing the loop from evaluation back to data conversion, policy learning, and deployment refinement.

\section{Competition Results}

Table~\ref{tab:leaderboard} reports normalized, weighted stage contributions, not raw scores. For the simulation, real-robot, and on-site stages, raw maxima \(M_i\) were \(300,300,50\), and weights \(w_i\) were \(0.20,0.30,0.50\), respectively. For raw score \(r_i\), the final score (out of 100) was
\begin{equation}
S_{\mathrm{final}}=100\sum_{i} w_i\frac{r_i}{M_i}.
\label{eq:final_score}
\end{equation}
Each stage entry is the corresponding term in~\eqref{eq:final_score}, with a maximum of 20, 30, or 50 points; the Final column sums them. NUS-CLEAR ranked first with 71.0 points, followed by RCL-Lab with 69.1 and DuanYanWuPing with 56.6.

The stage scores reveal different strengths across the competition. Youth2Real and mcg earned the two highest simulation contributions, while NUS-CLEAR and RCL-Lab led the final ranking through stronger combined real-robot and on-site performance. Deeptouch.ai earned the third-highest on-site score but placed sixth overall because of its lower earlier-stage contributions. Because the on-site final most directly tested deployment on the target hardware, the three highest-scoring on-site teams---NUS-CLEAR, RCL-Lab, and Deeptouch.ai---were invited to contribute the approach descriptions in the next section. A summary of the approaches adopted by each team is provided in Table~\ref{tab:cross_team}.

\begin{table*}[t]
\caption{Normalized, weighted stage contributions and final scores. Boldface identifies the three highest on-site entries, whose teams contributed the approach descriptions.}
\label{tab:leaderboard}
\centering
\small
\setlength{\tabcolsep}{3pt}
\begin{tabular*}{0.85\textwidth}{@{\extracolsep{\fill}}lrrrr@{}}
\toprule
Team & Simulation & Real robot & On-site & Final \\
\midrule
NUS-CLEAR & 16.7 & 20.0 & \textbf{34.3} & 71.0 \\
RCL-Lab & 15.5 & 20.6 & \textbf{33.0} & 69.1 \\
DuanYanWuPing & 16.9 & 15.0 & 24.7 & 56.6 \\
Youth2Real & 18.8 & 16.0 & 17.6 & 52.4 \\
mcg & 18.5 & 13.2 & 19.4 & 51.1 \\
Deeptouch.ai & 7.3 & 11.0 & \textbf{30.8} & 49.1 \\
\bottomrule
\end{tabular*}
\end{table*}

\begin{table*}[t]
\caption{Comparison of training and deployment details for the three contributed approaches in the on-site final. More details are provided in Section~\ref{sec:team_approaches}.}
\label{tab:cross_team}
\centering
\scriptsize
\setlength{\tabcolsep}{2.6pt}
\renewcommand{\arraystretch}{1.08}
\begin{tabular}{
>{\raggedright\arraybackslash}p{0.125\textwidth}
>{\raggedright\arraybackslash}p{0.205\textwidth}
>{\raggedright\arraybackslash}p{0.19\textwidth}
>{\raggedright\arraybackslash}p{0.205\textwidth}
>{\raggedright\arraybackslash}p{0.215\textwidth}}
\toprule
Team, policy, and task & Training steps and stage adaptation & Action and temporal interface & Learning rate and batch size & Data treatment and validation \\
\midrule
\textbf{NUS-CLEAR}: $\pi_{0.5}$; task-specific checkpoints
& Parts Righting: 10K; Bottle PnP: 6K; Package Scanning: 8K. The Stage~2 checkpoint initialized Stage~3 training.
& Delta actions; prediction/execution steps: 10/6. Observation history: 1.
& Fixed $5\!\times\!10^{-6}$. Batch: 32.
& Duration-quantile filtering; held-out outliers and active-dimension MSE for checkpoint selection; inactive-arm masking. \\
\midrule
\textbf{RCL-Lab}: GR00T N1.7; task-specific checkpoints
& Parts Righting: 20K; Bottle PnP: 25K; Package Scanning: 20K. The Stage~2 checkpoint initialized Stage~3 training.
& Delta actions; prediction/execution steps: 16/8. Observation history: 1.
& $5\!\times\!10^{-5}$; cosine schedule; warm-up ratio 0.05; weight decay $5\!\times\!10^{-6}$. Batch: 64.
& All released demonstrations; 50/75/100\% subset curriculum; no manual filtering; final checkpoint. \\
\midrule
\textbf{Deeptouch.ai}: $\pi_{0.5}$; Metal Parts Righting and Express Package Scanning
& Parts Righting: 8K; Package Scanning: 15K. The Stage~2 checkpoint initialized Stage~3 training.
& Absolute actions; prediction/execution steps: 50/10. Observation history: 1.
& Peak $2.5\!\times\!10^{-5}$; 1K warm-up; cosine decay to $2.5\!\times\!10^{-6}$ over 30K. Batch: 64.
& Duration/motion filtering; rare-flip subset and 2$\times$ oversampling; inactive-arm masking. \\
\addlinespace[2pt]
\textbf{Deeptouch.ai}: Diffusion Policy; Daily Chemical Bottle Pick-and-Place
& 625K steps, trained from scratch.
& Absolute actions; prediction/execution steps: 16/8. Observation history: 2.
& $1\!\times\!10^{-4}$; 500-step warm-up; cosine schedule. Batch: 256.
& All 1,000 demonstrations; random image cropping; selected epoch-970 checkpoint. \\
\bottomrule
\end{tabular}
\end{table*}

\section{Competition Teams and Approaches}
\label{sec:team_approaches}

\subsection{NUS-CLEAR}

\begin{figure}[t]
    \centering
    \includegraphics[width=0.98\linewidth]{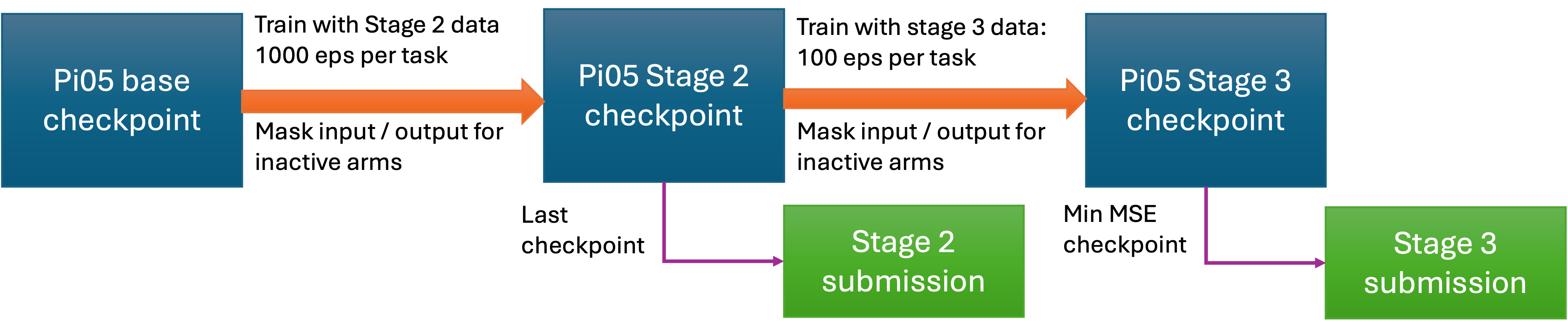}
    \caption{NUS-CLEAR's stage-wise adaptation pipeline. The pretrained $\pi_{0.5}$ base checkpoint was fine-tuned for Stage~2, submitted for Stage~2 evaluation, and then further adapted into the Stage~3 checkpoint used for the final submission.}
    \label{fig:team_a_pipeline}
\end{figure}

NUS-CLEAR used $\pi_{0.5}$ \cite{physicalintelligence2025pi05}, implemented with the \texttt{openpi}
codebase, as the base policy model and represented
robot actions as delta joint positions. The policy predicted 10-step action chunks and executed six steps from each prediction before replanning. During the real-robot
stage, NUS-CLEAR fine-tuned the pretrained model using the provided task
demonstrations. The on-site final used the same task definition as the
real-robot stage, but the physical domain was not identical: lighting
changed, object initial placements could vary across resets, and the
scene setup and calibration could shift slightly during robot motion
or competition preparation. NUS-CLEAR therefore treated the transition from the
real-robot stage to the on-site final as a continual adaptation problem across the fixed Stage~2 and Stage~3 data releases. The
policy was initialized from the checkpoint produced in the real-robot stage and then
continued to train offline on the fixed on-site dataset released for Stage~3. This
setting required the model to incorporate data from a shifted distribution
without degrading capabilities learned earlier. Figure~\ref{fig:team_a_pipeline}
summarizes this stage-wise checkpoint adaptation and submission pipeline.

The main difficulty was that the dataset from the on-site final was much smaller than
the dataset from the real-robot stage. For NUS-CLEAR, the model from the real-robot stage was trained with
approximately 1,000 episodes, while the on-site adaptation set contained
only on the order of one hundred trajectories. Direct fine-tuning on all
new trajectories can be sensitive to bad demonstrations, reset artifacts,
hesitation, repeated attempts, or rare execution patterns. The team therefore
first focused on data quality before model selection.

For trajectory-level data curation, the team used episode length as a simple
proxy for potential outliers. Let $\mathcal{D}_{\mathrm{on}} =
\{\tau_i\}_{i=1}^{N}$ be the on-site dataset and let $T_i$ denote the
length of trajectory $\tau_i$. The team computed empirical length quantiles
$Q_{\alpha}$ and kept only trajectories whose lengths fell within a
central range:
\begin{equation}
\mathcal{D}_{\mathrm{train}}
=
\{\tau_i \in \mathcal{D}_{\mathrm{on}}
\mid
Q_{\alpha_{\min}} \leq T_i \leq Q_{\alpha_{\max}}\}.
\end{equation}

In the implementation, $\alpha_{\min}=0.05$ and $\alpha_{\max}=0.95$ were used, corresponding to a conservative $5\%$ trimming of each tail of the trajectory-length distribution. Inspection of trajectories in these tails showed that unusually long episodes frequently contained hesitation, inefficient actions, or extended periods in undesirable states, whereas unusually short episodes were often associated with incomplete demonstrations, early failures, or reset artifacts. These trajectories were therefore excluded from gradient updates, as they provided less reliable targets for behavior cloning.

Long trajectories may also contain useful recovery behaviors. However, given the limited dataset size (approximately 100 episodes), such recovery segments were typically too sparse to compensate for the larger amount of undesirable behavior that preceded them. Including these trajectories could therefore expose the policy to failure-inducing actions without providing sufficient examples of how to recover from the resulting states. In this low-data regime, discarding such trajectories entirely was found to be more effective. A more principled alternative would be to annotate trajectory quality at a finer temporal granularity, for example over short segments of several frames, and use these annotations for advantage-weighted or quality-conditioned learning~\cite{intelligence2025pi}. Such fine-grained annotation, however, would require substantially more manual effort and was impractical within the limited duration of the challenge.

Importantly, the excluded trajectories were not discarded. Instead, they were retained as a held-out diagnostic set for checkpoint selection. The central-length trajectories provided relatively consistent targets for on-site adaptation, while the held-out outliers were used to detect excessive specialization to the small curated subset. The underlying objective was to adapt the policy to the new on-site environment without drifting too far from the capabilities inherited from Stage~2. In particular, the excluded trajectories could contain states and recovery behaviors that were underrepresented in the curated training set. Performance on these trajectories therefore served as a proxy for whether fine-tuning preserved behaviors acquired by the base policy rather than overwriting them through overfitting to the limited on-site data.

For each training epoch, the team sampled states from the held-out trajectories
and evaluated open-loop prediction of the next $K$ delta-joint actions.
For checkpoint $\theta_e$, the validation error was computed over the
active action dimensions as
\begin{equation}
\mathcal{E}(\theta_e)
=
\frac{1}{|\mathcal{S}|K|\mathcal{J}_{\mathrm{act}}|}
\sum_{(i,t)\in\mathcal{S}}
\sum_{h=0}^{K-1}
\sum_{j\in\mathcal{J}_{\mathrm{act}}}
\left(
\hat{a}_{i,t+h,j}^{\theta_e}
-
a_{i,t+h,j}
\right)^2 ,
\end{equation}
where $\mathcal{S}$ is the set of sampled validation states,
$\mathcal{J}_{\mathrm{act}}$ is the set of active action dimensions,
$a_{i,t+h,j}$ is the demonstrated delta-joint action, and
$\hat{a}_{i,t+h,j}^{\theta_e}$ is the policy prediction. NUS-CLEAR deployed
the checkpoint with the lowest held-out open-loop error. Although this
metric does not perfectly predict closed-loop robot success, it provided
a practical signal for detecting when continued fine-tuning began to
memorize the limited on-site training subset.

A second practical design choice was to remove unused state and action dimensions. Some final on-site scenes were effectively single-arm manipulation tasks, even though the robot interface exposed observations and commands for both arms. For such tasks, the RGB observation associated with the inactive arm, together with its proprioceptive state, was omitted from the policy input. The corresponding action dimensions were also removed from the training targets, so no loss was computed for joints belonging to the inactive arm.

Retaining the passive arm in the prediction space can be detrimental even when it is not required for task completion. Small prediction errors in unused joints may accumulate over long-horizon rollouts and produce visible drift. In a dual-arm workspace, such drift can alter the camera viewpoint, move the passive arm into the active workspace, increase the risk of self-collision, or inadvertently disturb nearby objects. Removing the corresponding observation and action dimensions therefore reduced unnecessary model complexity and prevented irrelevant prediction errors from affecting closed-loop execution, resulting in more stable deployment.

Overall, NUS-CLEAR's approach combined pretrained policy fine-tuning with
conservative on-site adaptation. The checkpoint from the real-robot stage provided a
strong initialization, the on-site trajectories supplied domain-specific
corrections, quantile-based filtering improved the quality of the
training data,
held-out length outliers provided an open-loop diagnostic, and action-space masking was intended to reduce passive-arm drift during deployment.

\subsection{RCL-Lab}

RCL-Lab used NVIDIA Isaac GR00T N1.7 \cite{nvidia2026gr00tn17} as the unified policy backbone for the final competition. GR00T N1.7 is a vision-language-action foundation model for humanoid manipulation that maps multimodal observations and language instructions to continuous robot actions. The team followed a task-specialized fine-tuning strategy: the same backbone, observation interface, and training recipe were used for all three scenes, but one separate checkpoint was trained for each task. This design avoided manual task-specific policy engineering while still allowing each model to specialize in the action distributions and visual contexts of Metal Parts Righting, Daily Chemical Bottle Pick-and-Place, and Express Package Scanning.

The action interface was intentionally kept close to the official Kuavo joint-control setting. The policy predicted delta actions in 16-step chunks and executed eight steps from each prediction across all tasks. The team did not add special treatment for end-effector commands, such as gripper-command binarization, delay compensation, thresholding, or separate smoothing. During real-robot deployment, RCL-Lab used the official inference and deployment scripts without modifying the control wrapper, recovery logic, or robot-side post-processing. This kept deployment aligned with the official baseline and concentrated the entry's engineering changes in backbone fine-tuning rather than custom control.

The team did not perform trajectory-level data filtering, manual demonstration cleaning, or task-specific data relabeling. All officially released demonstrations were kept in the training set. Instead of relying on data curation, the team's main adaptation mechanism was to continue training from the real-robot stage into the on-site final. The team first fine-tuned GR00T N1.7 on the Stage~2 real-robot dataset and then continued training from that checkpoint on the Stage~3 on-site release. Carrying the Stage~2 checkpoint forward preserved the earlier task representation while the Stage~3 data adapted the policy to the final robot, camera setup, object distribution, and reset conditions.

A practical component of the method was the progressive expansion of data during training.
For each task dataset $\mathcal{D}$, the team did not begin optimization with all demonstrations at once.
The team first trained on a $50\%$ subset, then resumed training after expanding the dataset to $75\%$, and finally continued training on the complete dataset.
Let $\mathcal{D}^{(p)}$ denote a $p$-fraction subset of the available data, with $p_k \in \{0.50, 0.75, 1.00\}$.
The training schedule can be summarized as

\begin{equation}
\theta_{k}=\mathrm{Train}(\theta_{k-1}, \mathcal{D}^{(p_k)}),
\qquad \theta_{0}=\theta_{\mathrm{GR00T}}.
\end{equation}

This procedure acted as a cold-start curriculum. The smaller initial subset helped the model adapt rapidly to the new task interface and dominant motion pattern, while later expansion exposed it to additional variation after the policy had already learned a stable initial behavior. In the team's experience, this gradual expansion was more manageable than starting immediately from the full dataset, especially under limited compute, limited submission opportunities, and no local real-robot validation.

Checkpoint selection was also constrained by the competition setting. Because the number of leaderboard submissions was limited and the team could not run local closed-loop robot tests, the team used the last saved checkpoint from each training run for evaluation rather than selecting checkpoints through real-robot rollouts or a separately designed validation set. This was a simple strategy and did not attempt to optimize a task-specific offline metric. It also exposed an important limitation of the entry: the team could not systematically compare whether lower validation loss, earlier checkpoints, or filtered data would have improved real-robot performance. Nevertheless, the approach remained competitive, suggesting that a strong pretrained VLA backbone combined with stable adaptation can be effective even with minimal deployment-side engineering.

Among the three final tasks, the team's policy performed strongest on Metal Parts Righting. The team believed this task benefited from the relatively structured object geometry and repeatable manipulation pattern, which matched the strengths of behavior cloning from fixed demonstrations. The most challenging task for the team's system was Express Package Scanning. Failures were mainly associated with grasping and transferring soft parcels. Compared with rigid metal parts or the bottle used in Daily Chemical Bottle Pick-and-Place, soft parcels deform under contact, provide less repeatable grasp geometry, and can slip or change shape during lifting and placement. Since the team's system did not include task-specific grasp correction, recovery, or package-orientation logic, these failures were difficult to correct once the policy deviated from the demonstrated state distribution.

Overall, the team's entry emphasized a simple but practical lesson for fixed-data real-robot learning. The team found that a small amount of on-site data can be more important than a much larger dataset from the real-robot stage when it captures the final hardware and scene distribution well. However, because the team trained on all released data without filtering, the value of this small on-site set was strongly quality-sensitive: inconsistent or noisy demonstrations could affect the final policy directly. The team's progressive training schedule provided a lightweight way to cold-start the model on a new task and then steadily incorporate more data. For teams with limited personnel, limited hardware access, and few evaluation submissions, this strategy offers a reproducible path for adapting a large humanoid foundation model to fixed-data industrial manipulation tasks.

\subsection{Deeptouch.ai}

Deeptouch.ai earned 30.8 points in the on-site final, ranking third at that stage, and finished sixth overall with a final score of 49.1.
The approach from Deeptouch.ai combined pretrained policy adaptation, task-dependent model
selection, and failure-driven data curation. The team used $\pi_{0.5}$ as the
base pretrained policy \cite{physicalintelligence2025pi05} and first fine-tuned it on the three task datasets
provided in the real-robot stage. The resulting checkpoint was
then used as the initialization for further on-site training on Metal Parts Righting
and Express Package Scanning, providing a domain-adapted starting point for the Kuavo robot
platform, competition camera views, and released action distributions.

The team selected policy classes according to task structure rather than using
a single model family throughout. Daily Chemical Bottle Pick-and-Place was the clearest example: the team's
system achieved the only full score on an individual task in the on-site
final using a comparatively small Diffusion Policy trained from scratch,
rather than a large pretrained VLA model. Before the final, the team observed
that Daily Chemical Bottle Pick-and-Place involved a dexterous-hand end effector, customized grasping
motions, bimanual handover, and multiple feasible approach and transfer
trajectories. These properties made the task only partially aligned with
the action and embodiment distributions likely covered by large VLA
pretraining, while making it well suited to a generative imitation policy
that can represent multimodal actions \cite{chi2023diffusionpolicy}. In
the team's experience, Diffusion Policy can appear weak if trained only for the
tens-of-thousands-step regime commonly sufficient for some behavior
cloning models. The selected policy was therefore trained from scratch for approximately 625K steps on Daily Chemical Bottle Pick-and-Place,
which gave a compact policy specialized to the dexterous-hand and
handover dynamics. This result suggests that for tasks with narrow but
nonstandard action structure, small task-specialized policies can be more
deployable within limited on-site time than larger general-purpose
models.

For longer-horizon tasks requiring stronger semantic and stage understanding, a training-time instruction decomposition was used. Demonstrations corresponding to different task stages or scene states were assigned more specific language instructions during training, providing additional supervision for learning distinct behaviors. At inference time, however, the policy received only the single general task instruction used by the evaluation protocol and had to infer the appropriate behavior directly from the visual observation.

Express Package Scanning followed this strategy. The correct action sequence depended on the observed package state: when the reverse side was facing upward, an additional flipping motion was required before scanning and transfer; otherwise, this step was unnecessary. During training, demonstrations from both states were first trained under the general test-time instruction. The data were then partitioned by package state and augmented with more specific instructions indicating the normal or reverse-side-up case. These state-specific descriptions provided a more structured training signal, while deployment still used only the general instruction, requiring the policy to infer the correct action branch from vision.

The reverse-side-up mode was selected through duration- and motion-based
filtering. Let $\mathcal{D}_3=\{\tau_i\}_{i=1}^{N}$ denote the Express Package Scanning
dataset, where each ROS-bag trajectory $\tau_i$ has duration $T_i$. The team
estimated $F_T(t)=N^{-1}\sum_{i=1}^{N}\mathbf{1}[T_i \leq t]$ and used
the median $q_{0.5}=\inf\{t:F_T(t)\geq 0.5\}$ as a threshold, based on
the prior that normal and reverse-side-up cases were approximately
balanced. Since flipping trajectories were expected to be longer, the team
first selected
\begin{equation}
\mathcal{D}_{\mathrm{flip}}^{T}
=
\{\tau_i\in\mathcal{D}_3 \mid T_i > q_{0.5}\},
\end{equation}
as candidate reverse-side-up demonstrations. Because slow normal
executions could also be long, the team further used features tied to the
additional flipping behavior. For each candidate trajectory, the team computed
the left tool-center-point (TCP) positions
$X_i^{L}=\{x_{i,s}^{L}\in\mathbb{R}^{3}\}_{s=1}^{n_i}$ in the robot base
frame. These positions were obtained by applying forward kinematics to
the arm joint angles to recover the final flange pose, and then composing
it with the flange-to-tool transform specified by the end-effector URDF.
The team also used the left gripper command sequence
$G_i^{L}=\{g_{i,s}^{L}\in\{0,1\}\}_{s=1}^{n_i}$, where the binary command
indicates gripper open or close. The reverse-side-up mode typically
introduced an extra gripper open--close event and additional modes in the
left 3-D TCP spatial distribution. These statistics formed the descriptor
$\phi(\tau_i)=(X_i^L,G_i^L)$, and the final filtered subset was
\begin{equation}
\mathcal{D}_{\mathrm{flip}}
=
\{\tau_i\in\mathcal{D}_{\mathrm{flip}}^{T}
\mid
\phi(\tau_i)\in\Omega_{\mathrm{flip}}\},
\end{equation}
where $\Omega_{\mathrm{flip}}$ denotes the empirical region corresponding
to the reverse-side-up mode in the left TCP-position and gripper-command
space. The selected trajectories were manually
checked, up-weighted, and trained for longer durations so that the policy
could better learn this rare but task-critical behavior.

The team also reduced irrelevant action dimensions for tasks that were
effectively single-arm manipulation. Although the robot interface exposed
commands for both arms, one arm could remain passive in some scenes.
Including passive-arm joints in the behavior-cloning target introduced
unnecessary input-output correlations and could slow learning under the
limited on-site budget. The team therefore applied both input and output masks:
passive-arm state dimensions were removed or zeroed when they were not
needed, and passive-arm action dimensions were masked at the policy
output so that the deployed controller did not execute irrelevant arm
motions. This reduced interference from robot degrees of freedom that did
not determine task success and made the learned policy easier to deploy
under limited on-site training time. Overall, the entry from Deeptouch.ai emphasized
practical offline-to-real adaptation: pretrained initialization where
useful, from-scratch task-specialized policies where the action
distribution demanded it, training-time instruction decomposition to structure state-dependent behaviors without modifying the inference interface, and targeted data filtering for rare failure
modes.

\section{Discussion}

\subsection{Generalization versus Specialization}

The transition from Stage 2 to the on-site final provides a revealing test of generalization. The task instructions, robot platform, and overall workcell structure were intentionally kept unchanged; the remaining shift arose mainly from naturally occurring variations such as lighting, background appearance, calibration, table geometry, and object poses. Nevertheless, NUS-CLEAR observed that directly deploying its Stage-2 checkpoint performed substantially worse than checkpoints continually adapted using the small on-site dataset. This suggests that even large pretrained VLA models can remain sensitive to modest deployment shifts, which is particularly relevant when policies trained in-house are transferred to factory sites.

Continual fine-tuning offers a practical remedy and was adopted by multiple teams, but it introduces a competing risk: a small adaptation set can drive the policy toward over-specialization and erase useful behaviors learned from the larger source dataset. NUS-CLEAR mitigated this through held-out open-loop evaluation for checkpoint selection. Future work should investigate stronger mechanisms for balancing adaptation and retention, including regularization that constrains policy drift during fine-tuning and more principled validation criteria for detecting when specialization begins to harm broader generalization.

\subsection{Data Quality versus Quantity}

The competition highlights that demonstration quality can be as important as quantity, especially when adapting a pretrained policy with a small amount of task-specific data. Several teams therefore filtered or reweighted demonstrations rather than treating all trajectories equally. NUS-CLEAR, for example, used episode length to remove unusually short or long trajectories, while Deeptouch.ai used task-specific motion statistics to identify rare but important behaviors.

However, episode-level curation is inherently coarse. A problematic trajectory may contain both failure-inducing actions and valuable recovery behavior. Keeping the entire episode teaches the policy both, whereas removing it discards useful recovery examples. This trade-off is particularly unfavorable in small adaptation datasets. As observed by NUS-CLEAR, undesirable behaviors may occupy a substantial portion of an imperfect trajectory, while recovery segments are comparatively sparse. The policy may therefore learn failure-inducing actions more easily than robust recovery, increasing the likelihood of entering poorly represented states during deployment. In such cases, the negative effect of retaining the trajectory can outweigh the benefit of its recovery examples.

This motivates finer-grained, \emph{segment-level} data curation. Individual segments could be identified as successful progress, failure-inducing behavior, or recovery, allowing harmful portions to be removed while retaining useful corrective actions. Learned value or quality functions could further automate this process and support advantage-weighted or quality-conditioned learning. More generally, developing reliable segment-level measures of demonstration quality may be more effective than relying on coarse episode-level heuristics such as trajectory duration~\cite{intelligence2025pi}.

\subsection{Failure Analysis and Recovery}

A consistent observation from the competition was that Express Package Scanning was the most challenging task, with generally the lowest scores across participating teams. A major source of failure was the manipulation of soft parcels, particularly during flipping. Unlike rigid objects, parcel motion depends strongly on deformation, friction, contact location, and grasp configuration. Small variations can therefore cause a nominal flipping action to stop halfway, slip, or leave the parcel in an intermediate configuration that is rarely represented in the demonstrations.

Such states expose a key weakness of behavior-cloned policies. Once execution enters an out-of-distribution state, the predicted action becomes less reliable and may move the system even further away from the demonstrated trajectories, causing errors to accumulate until the task fails. This suggests that robustness depends not only on accurately reproducing nominal demonstrations, but also on learning how to recover from unexpected intermediate states. More systematic recovery data could therefore be introduced through segment-level curation, DAgger-style corrective data collection, or reinforcement-learning-based post-training.

\subsection{Evaluation Matters}

Training a robot policy involves many choices, including data selection, training duration, hyperparameters, and checkpoint selection, making efficient evaluation critical. However, identifying which checkpoint will actually perform best on the real robot remains an open problem. NUS-CLEAR used held-out open-loop action prediction as a practical proxy for checkpoint selection and found it useful for detecting overfitting during on-site adaptation. Nevertheless, open-loop error does not fully capture closed-loop effects such as compounding errors, contact dynamics, or recovery ability, and therefore does not always correlate perfectly with real-robot success.

Direct hardware evaluation provides the most reliable signal but is highly inefficient. During the competition, each team had only a few real-robot evaluation opportunities per day, with only a limited number of physical rollouts available in total. Each evaluation also required robot setup, resetting objects and environments, and handling occasional hardware or system issues. Consequently, exhaustively comparing checkpoints on hardware was impractical. Developing offline or lightweight evaluation metrics that reliably predict closed-loop deployment performance therefore remains an important research problem for practical robot learning.

\subsection{Masking Irrelevant Observation and Action Dimensions}

Both NUS-CLEAR and Deeptouch.ai found it beneficial to remove observation and action dimensions associated with inactive robot components. While masking can reduce model complexity and improve training efficiency, a more important benefit is avoiding unnecessary distribution shift.

For example, in an effectively single-arm task, the inactive arm remains nearly stationary throughout the demonstrations. The policy therefore observes only a very narrow distribution of states for that arm. If the inactive arm is slightly displaced during deployment because of initialization error, control noise, or external disturbance, the policy can immediately encounter an observation configuration that is poorly represented in the training data. Predictions from this out-of-distribution state may then affect even the active arm, causing errors to compound.

NUS-CLEAR observed this effect particularly clearly during Stage~1 simulation experiments: when inactive-arm dimensions were retained, small perturbations of the unused arm could eventually destabilize the entire policy and, in some cases, cause the robot to collapse. Masking irrelevant inputs and outputs therefore does more than simplify learning; it removes unnecessary sources of distribution shift and can substantially improve closed-loop robustness.

\subsection{Model Scale Should Match Environmental Variation}

When data quantity and quality were broadly similar, frontier VLA backbones performed comparably: the $\pi_{0.5}$~\cite{physicalintelligence2025pi05}-based NUS-CLEAR entry and the GR00T N1.7~\cite{nvidia2026gr00tn17}-based RCL-Lab entry finished only 1.9 points apart. This single competition does not establish model equivalence, but it suggests that once backbone capability reaches the frontier, data adaptation, validation, and deployment details can matter as much as the choice between leading models.

Model size also interacts with environmental diversity. In simulation or other low-variation settings, a small task-specific policy such as Diffusion Policy~\cite{chi2023diffusionpolicy} can fit the environment more easily and may outperform a larger generalist model \cite{chi2023diffusionpolicy}. Physical deployment contains broader visual, geometric, contact, and hardware variation, so large foundation models generally benefit from stronger pretrained representations and greater generalization capacity. This is a tendency rather than a universal rule: the specialized Diffusion Policy used by Deeptouch.ai achieved a full score on Daily Chemical Bottle Pick-and-Place, showing that a narrow action distribution can still favor a compact specialist.

\section{Conclusion}

The \reali{} Challenge suggests that improving real-world robot learning is not simply a matter of training a larger policy on more demonstrations. Under a fixed data budget, performance is determined by how effectively a system extracts useful supervision from the available trajectories, adapts to the deployment distribution, selects among candidate policies, and prevents small prediction errors from becoming closed-loop failures. Across the leading teams, seemingly simple decisions---filtering demonstrations, retaining prior-stage knowledge, masking irrelevant dimensions, choosing task-appropriate policy classes, and selecting checkpoints carefully---had substantial consequences on the physical robot. Taken together, these results suggest that real-world policy learning should be viewed less as an isolated model-fitting problem and more as an integrated problem of \emph{data, adaptation, evaluation, and deployment}.

Perhaps the most important unresolved issue is evaluation. Real-robot rollouts provide the evidence that ultimately matters, yet they are too costly and sparse to guide the many decisions involved in policy development. Conversely, convenient offline metrics such as action-prediction error only imperfectly capture compounding errors, contact interactions, distribution shift, and recovery. Closing this gap may therefore be as important as developing the next policy architecture: practical embodied AI will require evaluation methods that can efficiently predict which data, checkpoint, adaptation strategy, or model will actually succeed after deployment.

\begingroup
\small
\setlength{\bibsep}{1pt plus 0.3pt}
\bibliographystyle{unsrtnat}
\bibliography{reali_ram_refs_v2,arxiv_refs}
\endgroup

\end{document}